\documentclass[letterpaper, 10 pt, conference]{ieeeconf}  

\IEEEoverridecommandlockouts                              

\usepackage{multirow}
\usepackage{amsmath}  
\usepackage{amssymb}  
\usepackage{color,array}
\usepackage{xcolor}
\usepackage{graphicx}
\usepackage{setspace}  
\usepackage{makecell}
\usepackage{subcaption}
\usepackage{caption}
\DeclareCaptionLabelSeparator{periodspace}{.\quad}
\usepackage{graphics} 
\usepackage{epsfig} 
\usepackage{times} 
\usepackage{amsmath} 
\usepackage{amssymb}  
\usepackage[linesnumbered,ruled,vlined]{algorithm2e}
\usepackage{subfloat}

\usepackage{booktabs} 
\usepackage{gensymb}  
\usepackage{bm}  
\usepackage{underscore}  

\usepackage{amsmath} 
\usepackage{amssymb}  

\usepackage{threeparttable}
\usepackage{enumerate}
\usepackage{bbm}

\usepackage{bbding} 
\usepackage{cite}
\usepackage{hyperref}
\usepackage{hhline}
\usepackage{colortbl}

\hypersetup{
citecolor = blue,
linkcolor = blue,
anchorcolor = blue,
urlcolor = blue
}
\author{Hanjing Ye$^{1}$, Jieting Zhao$^{1}$, Yu Zhan$^{1}$, Weinan Chen$^{2}$, Li He$^{1}$ and Hong Zhang$^{1*}$, \textit{Fellow IEEE} 
\thanks{$^{1}$Hanjing Ye, Jieting Zhao, Yu Zhan, Li He and Hong Zhang are with Shenzhen Key Laboratory of Robotics and Computer Vision, Southern University of Science and Technology. $^{2}$Weinan Chen is with Guangdong University of Technology. $^{*}$corresponding author (hzhang@sustech.edu.cn).}%
\thanks{Our code, video and appendix are available at \url{https://sites.google.com/view/oclrpf}.}%
}

\begin{document}
\title{\LARGE \bf
Person Re-Identification for Robot Person Following \\with Online Continual Learning}

\maketitle
\thispagestyle{empty}
\pagestyle{empty}

\begin{abstract}
        Robot person following (RPF) is a crucial capability in human-robot interaction (HRI) applications, allowing a robot to persistently follow a designated person. In practical RPF scenarios, the person can often be occluded by other objects or people. Consequently, it is necessary to re-identify the person when he/she reappears within the robot's field of view. Previous person re-identification (ReID) approaches to person following rely on a fixed feature extractor. Such an approach often fails to generalize to different viewpoints and lighting conditions in practical RPF environments. In other words, it suffers from the so-called domain shift problem where it cannot re-identify the person when his re-appearance is out of the domain modeled by the fixed feature extractor. To mitigate this problem, we propose a ReID framework for RPF where we use a feature extractor that is optimized online with both short-term and long-term experiences (i.e., recently and previously observed samples during RPF) using the online continual learning (OCL) framework. The long-term experiences are maintained by a memory manager to enable OCL to update the feature extractor. Our experiments demonstrate that even in the presence of severe appearance changes and distractions from visually similar people, the proposed method can still re-identify the person more accurately than the state-of-the-art methods.
\end{abstract}


\section{INTRODUCTION}

Robot person following (RPF) \cite{islam2019person} serves as an essential function in many HRI applications, enabling a robot to follow a specified person autonomously. However, the person being followed may become occluded in various situations, such as when other objects or people obstruct the view of the robot in the working environment. Therefore, it is crucial to re-identify the person when he re-appears in the view. 

Existing RPF systems can be achieved through two steps: \textit{identify} and \textit{follow}.  In the \textit{identify} step, the system performs tracking and possibly ReID to locate the target person, while the \textit{follow} step involves planning and executing the control of the robot to maintain the desired relative position with the target person.
In this paper, we focus on the ReID aspect, specifically re-identifying the target person after occlusion. Existing ReID methods for RPF describe a person's appearance either with hand-crafted features\cite{wang2022tim,koide2016identification} or with learned features\cite{koide2020monocular}. However, these methods may experience poor generalization when the features are not sufficiently discriminative for re-identifying the person. Some methods \cite{chen2017integrating,chen2016lbp} update the tracker online with newly acquired observations of the target person to distinguish the person from the background and other distracting individuals. Such solutions usually do not consider the appearance of a person explicitly, leading to suboptimal ReID performance. To improve the generalization ability, one possible solution is to train the feature extractor online with the most recently observed samples, i.e., short-term experiences. However, we found this would result in limited discriminative ability when the re-appearance of the target person is out of the learned domain represented by the short-term experiences. All these problems are commonly known as domain drift\cite{mai2022online}.

\begin{figure}[t]
    \centering
    \includegraphics[width=0.80\linewidth,trim=40 35 40 60,clip]{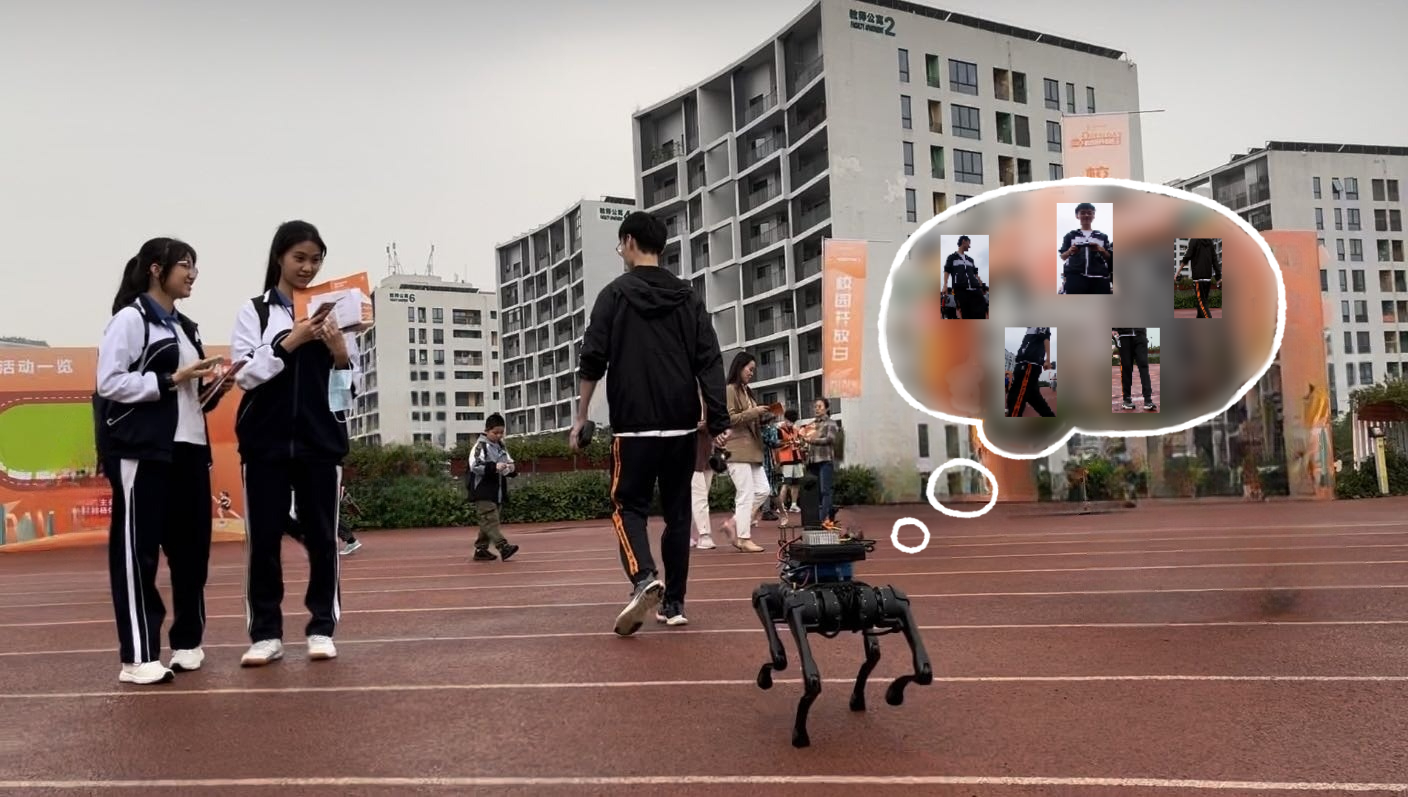}
    \caption{Robot person following with online continual learning. To this end, long-term and short-term experiences are utilized to optimize the feature extractor online to represent the discriminative appearance of the target person.}
    \label{introduction}
    \vspace*{-0.2in}
\end{figure}

To solve the above problems, we propose to utilize long-term experiences in addition to short-term ones to optimize the feature extractor online for representing the target person's discriminative appearance. Specifically, we approach the person ReID in RPF as a problem of online continual learning (OCL)\cite{mai2022online}, which aims to learn the newest knowledge without forgetting long-term experiences using a size-limited long-term memory.
This idea has shown promising results in existing works on dense mapping\cite{sucar2021imap} and place recognition\cite{yin2022bioslam}.
For example, IMap \cite{sucar2021imap} incrementally learns a NeRF-based dense map by replaying images and poses from a sparse keyframe set, where camera poses are estimated through the tracking process. Similarly, BioSLAM \cite{yin2022bioslam} constructs a discriminative long-term memory to replay point clouds and positions for learning a life-long place recognition network, where positions are obtained via LiDAR odometry.

To develop a ReID framework for RPF that capitalizes on long-term experiences, we established a long-term memory module designed to archive key historical samples, chosen via a loss-guided keyframe selection method. By integrating these long-term samples with short-term data, we optimize the feature extractor online to maintain a comprehensive understanding of the target person, bridging past and present knowledge. Additionally, we apply these optimized features to train a ridge-regression-based classifier for accurate target recognition.
Lastly, a ReID lifecycle management is implemented to form a complete ReID solution. In our experiments, the RPF system with our ReID method can reliably re-identify and follow the target person even in situations with visually similar distracting people and different appearances after occlusion.

\section{Related Work}

\subsection{Person ReID in Robot Person Following} \label{related-A}
Person ReID is crucial for RPF, which helps re-identify the target person after occlusion. Existing ReID methods in RPF usually describe the appearance of the target person with hand-crafted features\cite{wang2022tim,koide2016identification} or learned features\cite{koide2020monocular}. Examples of hand-crafted features include geometric attributes\cite{wang2022tim}, and characteristics like height, gait and clothing color\cite{koide2016identification}. Alternatively, ReID can rely on features learned from a ReID dataset. For example, \cite{koide2020monocular} trains a convolutional neural network (CNN) using a custom-built, small-scale ReID dataset and then extracts features from the low-level response maps of the CNN. Often, these features are further utilized to construct a target classifier with short-term experiences.

The above methods, however, often fail to re-identify the person in complicated RPF situations because features from a fixed feature extractor have a limited capability to generalize to different viewpoints and lighting conditions in practical RPF environments. To mitigate this generalization problem, we optimize the feature extractor online with the short-term experiences used to construct the target classifier in the above methods. However, optimized features are still not discriminative enough to recognize the target person, especially when the re-appearance of the target person is out of the learned domain. These are commonly referred to as domain drift\cite{mai2022online}. To mitigate these problems, we propose to utilize OCL techniques to collect valuable long-term experiences. These experiences, in addition to short-term ones, are used to optimize ReID features, thereby improving the ReID performance of the RPF system.

\subsection{Person ReID in Computer Vision}
Person ReID has been a prominent research area in computer vision, primarily identifying individuals in video surveillance systems\cite{leng2019survey}. 
Various methods have been proposed to solve the ReID problem. For instance, \cite{gray2008viewpoint} introduces a hand-crafted feature that combines eight color channels (RGB, HSV, and YCbCr) and 19 texture channels to achieve viewpoint invariance. Another approach\cite{Shi2015CVPR} involves using attribute-based features to achieve competitive ReID performance. However, in recent years, with the advancement of deep learning techniques, learned features\cite{ye2021deep} have become dominant in ReID research due to their end-to-end nature and excellent generalization. Notably, \cite{Li2014CVPR} proposes a CNN-based ReID method that effectively models complex photometric and geometric transformations. However, ReID with a global CNN feature can introduce distractive information in case of occlusion, posing a challenge in real-world scenarios.

To address the issue of occlusion, researchers have introduced ReID methods\cite{somers2023body,li2021diverse} that leverage pre-defined or learned part masks to match features defined with respect to parts of a target person. Considering that an occluded human body is frequently encountered in RPF scenarios, one can use part-guided ReID features to describe a person's appearance. Still, as mentioned before, these features from a fixed feature extractor have a limited generalization ability in practical RPF environments with different viewpoints and lighting conditions. To solve the generalization problem, a similar approach to ours is the memory-based ReID\cite{zhong2019invariance, zhong2020learning, zhao2021learning}, which tackles unsupervised domain adaptation by transferring knowledge from a labeled source domain to an unlabeled target domain. However, deploying these methods poses significant challenges in the context of RPF due to the demands for extensive iterative training and substantial memory storage. In contrast, our approach enables the optimization of the feature extractor and the re-identification of the target person in real time, even on onboard devices. 
In this paper, we are the first to explore pre-trained deep features from the computer vision community for forming a complete RPF-task-driven target-ReID framework.

\begin{figure*}[t]
    \centering
    \includegraphics[width=0.9\linewidth]{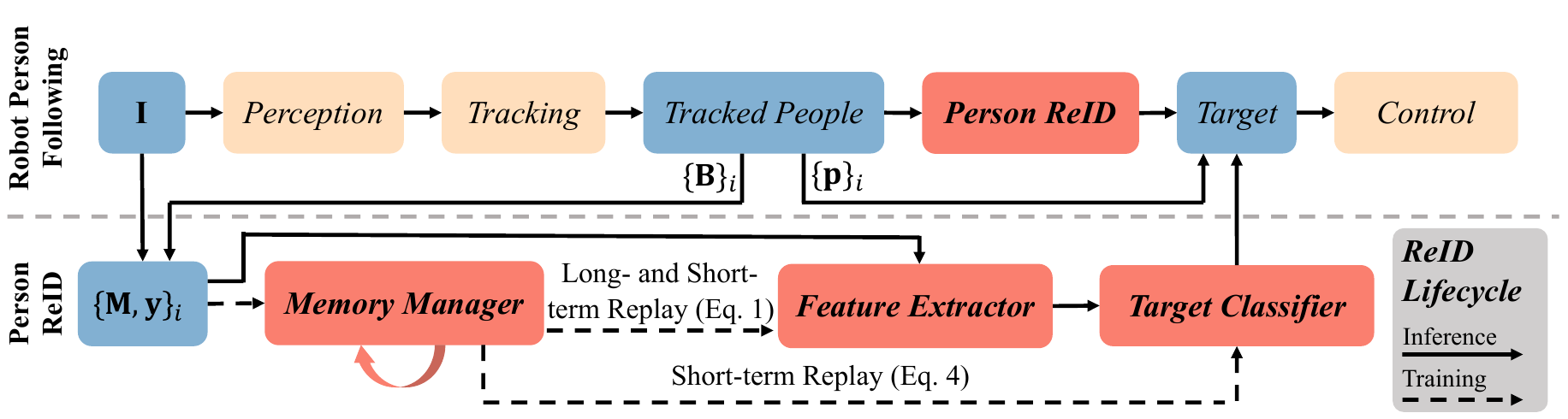}
    \caption{The top part is the pipeline of our RPF system and the bottom part is the proposed person ReID framework. 
    We obtain image patches $\{\mathbf{M}\}_i$ of the tracked people using the current image $\mathbf{I}$ and their bounding boxes $\{\mathbf{B}\}_i$. 
    When the target person is consistently tracked, his label $\mathbf{y}$ represents positives and other people are negatives. 
    Afterward, we add $\{\mathbf{M}, \mathbf{y}\}_i$ to the \textit{memory manager} for memorization. 
    Additionally, these patches are fed into the \textit{feature extractor} to extract ReID features. 
    These features are utilized by the \textit{target classifier} to estimate the target confidence. 
    If the target confidence is greater than a threshold, the corresponding position $\mathbf{p}$ is designated as the target position. 
    In addition to the inference above process, the \textit{memory manager} simultaneously replays long-term and short-term experiences to train the \textit{feature extractor}. Meanwhile, the \textit{target classifier} is trained with short-term experiences. 
    If the target person is not found among the tracked individuals, the training process pauses, and all observations $\{\mathbf{M}, \mathbf{y}\}_i$ become candidates for re-identification. 
    The above training and inference processes are managed by the \textit{ReID lifecycle}.
    }
    \label{method}
    \vspace*{-0.20in}
\end{figure*}

\subsection{Online Continual Learning}
OCL addresses the challenge of learning from a non-independent and identically distributed (Non-IID) stream of data in an online manner, with the objective of preserving and extending historical knowledge\cite{mai2022online}. The Non-IID data setting aligns with the observation scenario of our RPF system, in which the appearance of an observed individual significantly varies due to complex backgrounds and the motion of the robot and target.

Recent works in OCL can be categorized into three main families: regularization-based, parameter-isolation-based and memory-replay-based methods. Regularization-based methods\cite{kirkpatrick2017overcoming, schwarz2018progress} preserve knowledge by adding history-related constraints to the loss function during current task training, thereby balancing the loss gradient direction for old and new knowledge. However, these methods face challenges in finding the desired global optima, making it difficult to strike a balance between both types of knowledge. Parameter-isolation-based methods\cite{mallya2018packnet, leeneural} retain old knowledge by freezing the related parts of the model and only allowing the remaining parts to learn new knowledge. However, these methods are limited by the initial model capacity and require significant training time to achieve good performance. Memory-replay-based methods\cite{Rahaf2019nips, shim2021online,ER} utilize memory replays to learn old knowledge incrementally. Examples include Reservoir\cite{ER}, which randomly forgets samples based on a distribution related to observation times, MIR\cite{Rahaf2019nips}, which randomly updates the memory and retrieves ``the hardest'' samples for model updating, and ASER\cite{shim2021online}, which utilizes an Adversarial Shapley value scoring method for memory retrieval to preserve latent decision boundaries for previously observed samples.

Recently, the benefits of memory-replay-based OCL have been demonstrated in several works\cite{sucar2021imap,yin2022bioslam} to enhance the perception ability of robot systems. Therefore, we adopt a memory-replay-based algorithm in the implementation of our RPF system, although our solution is not limited to any particular OCL algorithm. To the best of our knowledge, we are the first to integrate the OCL concept into an RPF system to optimize the feature extractor incrementally from both long-term and short-term experiences.

\section{Method}

\subsection{Problem Statement and Overview}\label{sec3-1}

Our RPF system is an extension of our previous work\cite{ye2023icra}, represented by the top half of Fig.~\ref{method}. Our previous RPF system allows for accurate tracking of individuals, even in scenarios with partial occlusion. It first tracks multiple people and then identifies the target person to follow by selecting the corresponding identity (ID). However, when the target person undergoes occlusion and disappears from the camera view, his ID may be removed because no observation is associated with the ID. Therefore, re-identifying the target person after occlusion, either momentarily or over a long time, becomes crucial. To solve this problem in our current work, we introduce a person ReID process, which is performed by the module in the lower half of Fig.~\ref{method}.
In this ReID module, the \textit{feature extractor} and the \textit{target classifier} are optimized when the target person can be correctly identified from tracked people. Later, if and when a long-time occlusion occurs, the optimized models are utilized to re-identify the target person among all the tracked people.

In each ReID period, we capture image patches $\{\mathbf{M}\}_i$ of the tracked individuals using the current image $\mathbf{I}$ and their corresponding bounding boxes $\{\mathbf{B}\}_i$. When the target person is consistently tracked, his label $\mathbf{y}$ represents a positive sample, while labels for other people are negatives. Subsequently, these patches $\{\mathbf{M}\}_i$ are fed into the \textit{feature extractor} for extracting ReID features (Sec.~\ref{sec3-2}) and these features are further utilized by the \textit{target classifier} to estimate the target confidence (Sec.~\ref{sec3-3}). If the target confidence is greater than a threshold, the corresponding position $\mathbf{p}$ is designated as the target position.

In addition to the inference process mentioned above, we add $\{\mathbf{M}, \mathbf{y}\}_i$ to the \textit{memory manager} (Sec.~\ref{sec3-4}) for performing memory-replay-based OCL.
Specifically, the \textit{feature extractor} is incrementally optimized with both long-term and short-term experiences ($m_{\rm lt}\cup m_{\rm st}$) in an OCL manner through Eq.~\ref{eq1}.
Besides, the \textit{target classifier} is trained with short-term experiences $m_{\rm st}$ through Eq.~\ref{eq4}. If the target person is not found among the tracked people, the training process pauses, and all observations $\{\mathbf{M}, \mathbf{y}\}_i$ become candidates for re-identification. The above training and inference processes are managed by the \textit{ReID lifecycle} detailed in Algorithm I in APPENDIX-D. Except for the memory-replay-based OCL, we name this ReID framework as \textbf{RPF-ReID}, which is a complete \underline{RPF}-task-driven target-\underline{ReID} module based on pre-trained deep features from the computer vision community.

\subsection{Feature Extractor}\label{sec3-2}
We use a feature-based neural network to extract a person's appearance features. Given an image $\mathbf{I}$ and a person's bounding box $\mathbf{B}$, we extract his image patch, denoted as $\mathbf{M}$. Subsequently, we fine-tune a feature extractor $f$, a ResNet pre-trained on MOT16 ReID\cite{milan2016mot16}, which extracts local features from $\mathbf{M}$. To represent partially visible human bodies, we further transform these local features into features associated with the body parts \cite{ye2023icra}. These features are denoted as $\mathbf{F}\in{\mathbb{R}^{N\times C}}$, where $N$ represents the number of body parts and $C$ is the size of the feature dimension. Besides, a visibility indicator $v_i$ with $i\in \{1,\cdots,N\}$ is defined and set to $1$ if the $i_{th}$ body part is visible and $0$ otherwise.

\begin{figure}[t]
    \centering
    \begin{subfigure}[]{0.23\linewidth}
        \centering
        \includegraphics[width=\linewidth]{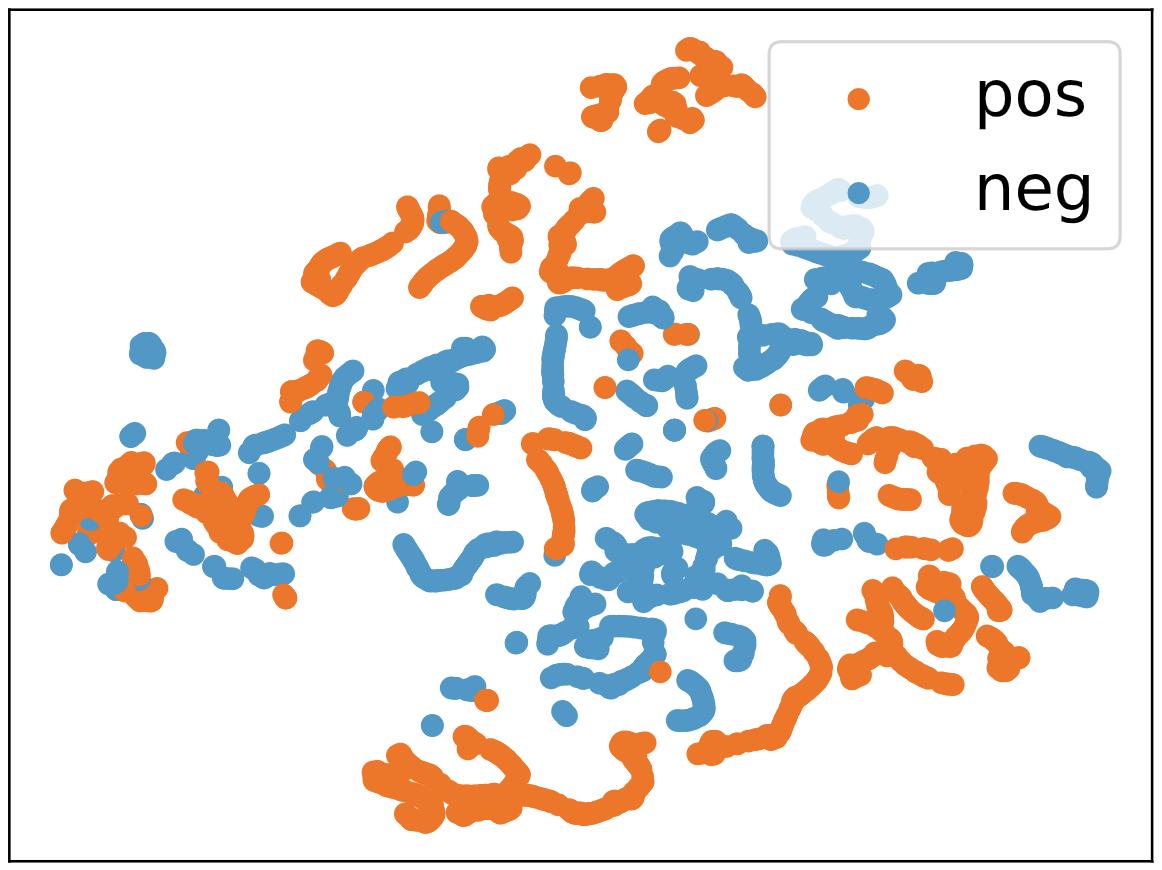}
        \caption{W/o S\&L}
        \label{tsne1}
    \end{subfigure}%
    \hspace{0.001\linewidth}
    \begin{subfigure}[]{0.23\linewidth}
        \centering
        \includegraphics[width=\linewidth]{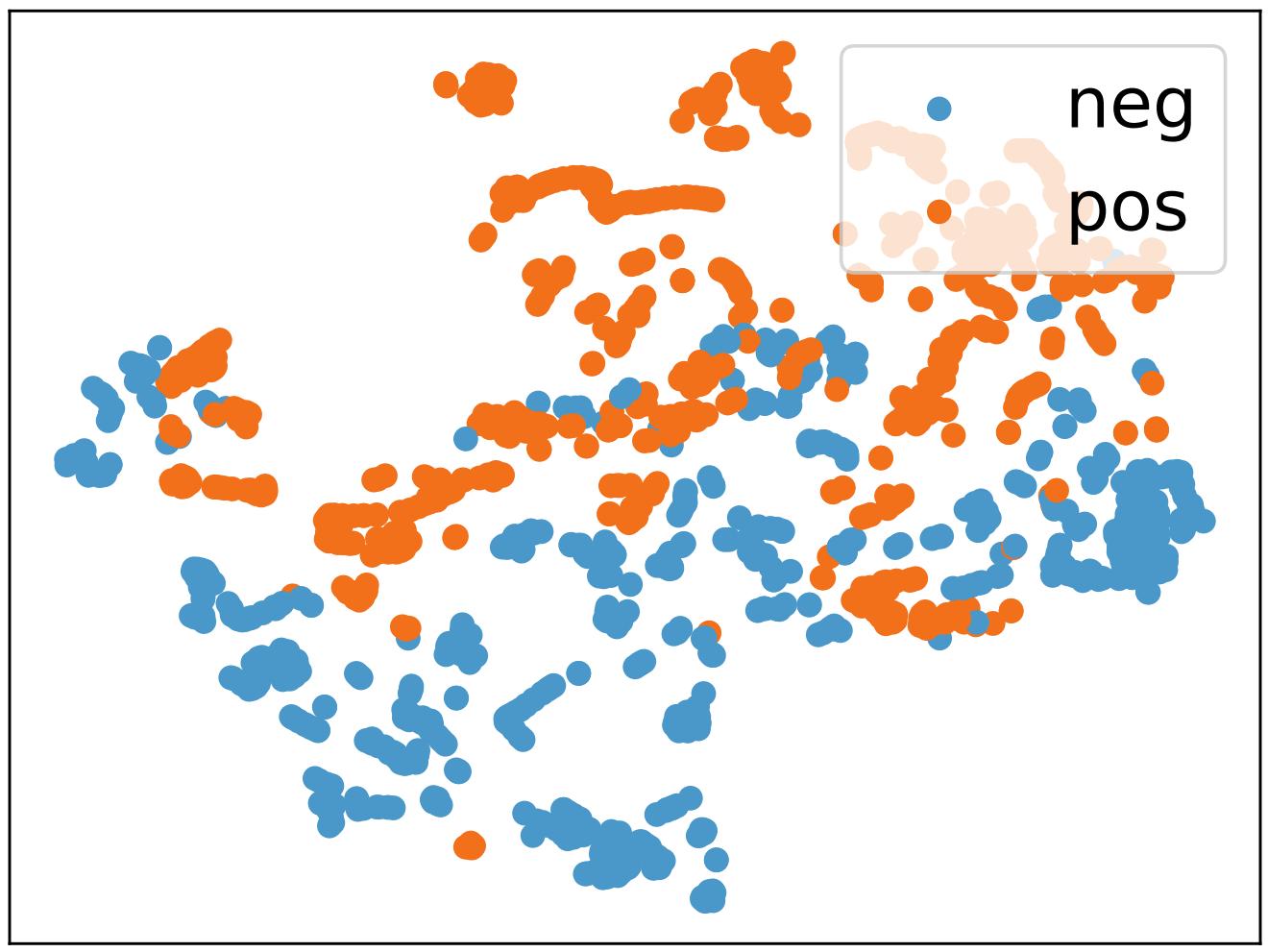}
        \caption{W/ S only}
        \label{tsne2}
    \end{subfigure}%
    \hspace{0.001\linewidth}
    \begin{subfigure}[]{0.23\linewidth}
        \centering
        \includegraphics[width=\linewidth]{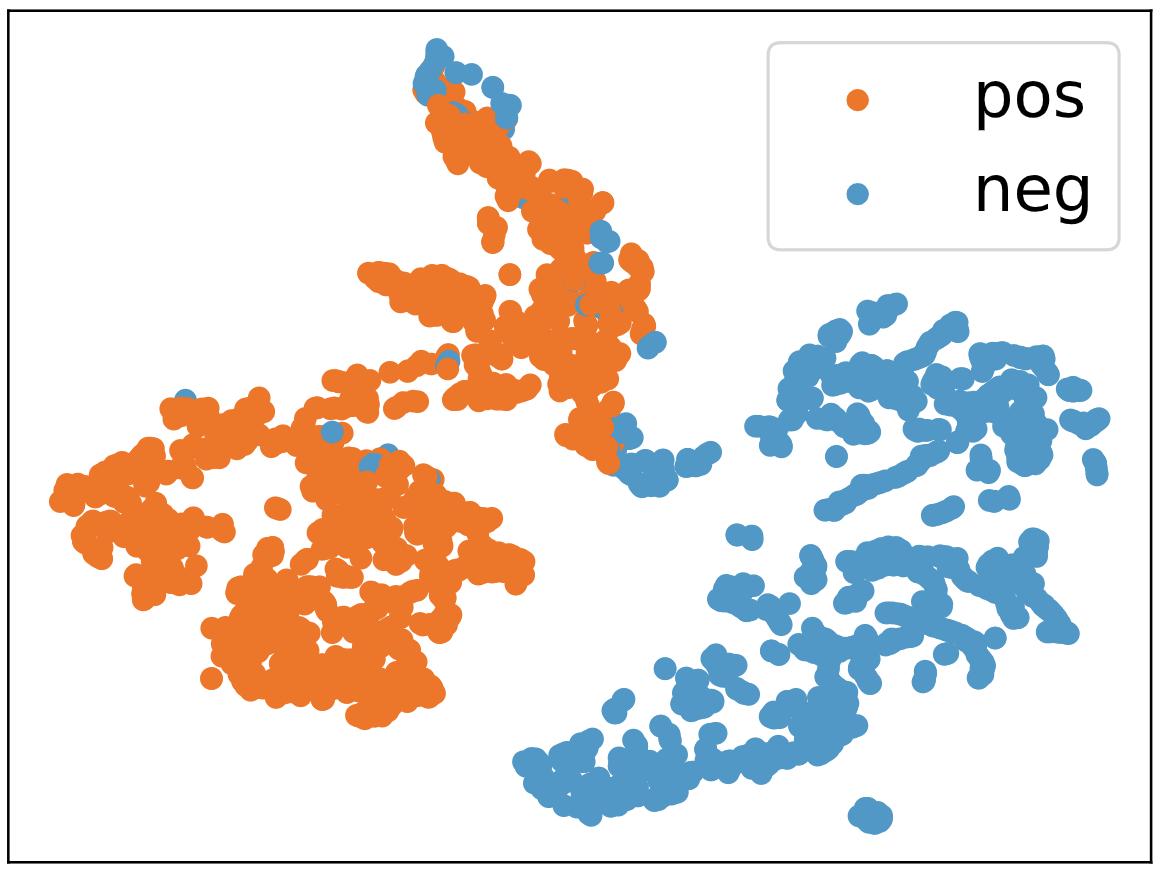}
        \caption{W/ S\&L}
        \label{tsne3}
    \end{subfigure}%
    \hspace{0.001\linewidth}
    \begin{subfigure}[]{0.23\linewidth}
        \centering
        \includegraphics[width=\linewidth]{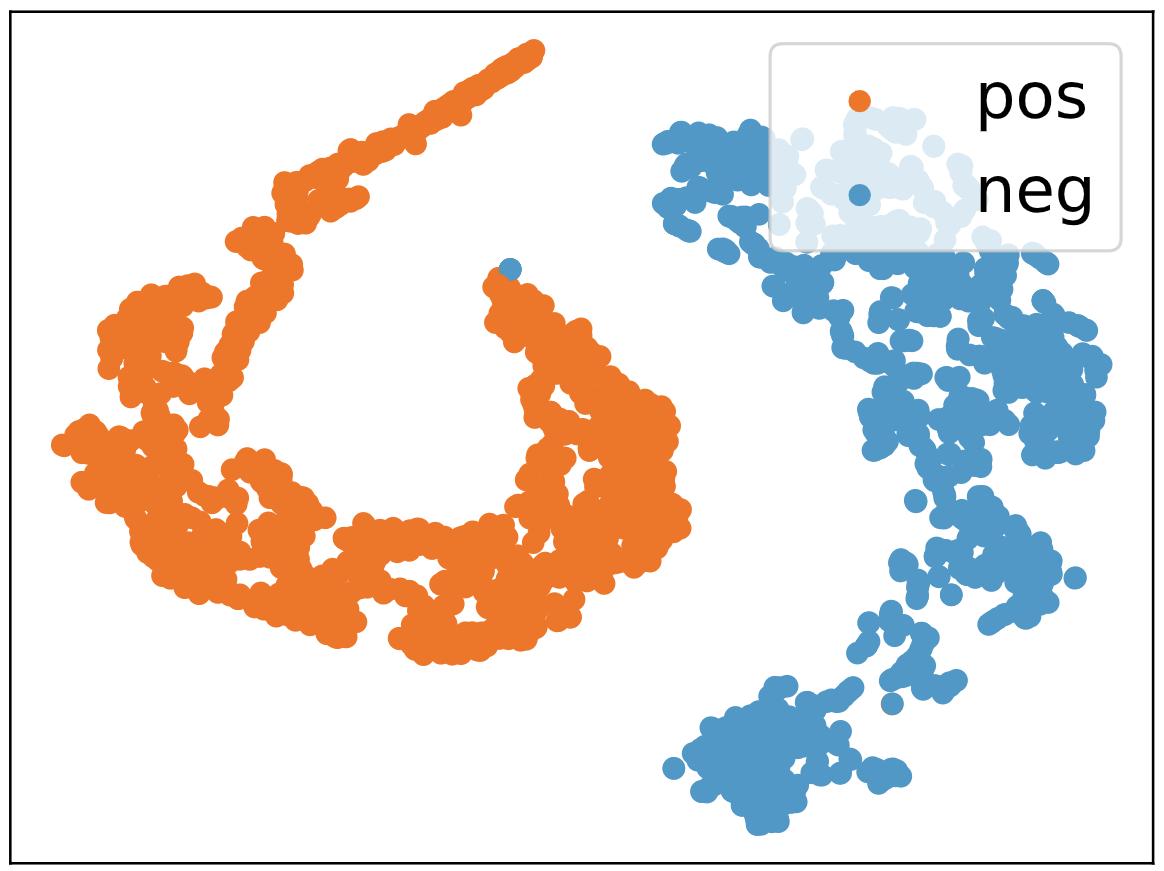}
        \caption{Ideal}
        \label{tsne4}
    \end{subfigure}%
\caption{Feature distribution of the target person (positive) and other distracting people (negative) across all observed samples at the end of the sequence. ``S'' represents short-term experiences and ``L'' for long-term ones. (a) Pre-trained features without any online optimization. (b) Trained features with online optimization using short-term experiences only. (c) Trained features using both short-term and long-term experiences within our framework. (d) Ideal feature distribution where features are optimized offline through extensive iterative training.}
\label{tsne}
\vspace*{-0.20in}
\end{figure}

In previous RPF works\cite{wang2022tim,koide2016identification,koide2020monocular}, the feature extractor is trained offline and fixed under the assumption of independent and identically distributed (IID) observations, i.e., the training and testing scenarios are assumed to be IID. However, this assumption may not be valid in an application such as our RPF. For instance, it may not hold when the target person's appearance is non-discriminative in the pre-defined feature space. One possible solution is to utilize short-term experiences to fine-tune the feature extractor online. However, optimized features are still not discriminative enough to recognize the target person.
These two problems are commonly referred to as domain drift\cite{mai2022online} and can be observed from Fig.~\ref{tsne} (a) and (b), respectively. Due to domain drift problems, the resulting ReID features fail to distinguish the target person from others across the observed samples in the sequence.

To address these problems, we adopt the concept of OCL \cite{mai2022online}. Instead of utilizing a fixed feature extractor, we continually fine-tune the feature extractor with both long-term and short-term experiences. Due to the requirement of efficient learning, OCL demands that the model is trained with only one limited batch at a time, and other batches are not included. In addition, OCL requires that the batch should contain current and historical samples. Therefore, we typically maintain a long-term memory, denoted as $\mathbb{L}$, to store a subset of historical samples. In every ReID period, $\mathbb{L}$ replays only one batch, denoted as $m_{\rm lt}\subset \mathbb{L}$. Besides, the most recent observed $K$ samples, denoted as $m_{\rm st}$, are included to represent the current knowledge. Our OCL formulation thus can be represented as follows:
\begin{equation}
    \mathop{\arg\min}_{\theta_f} \sum_{\substack{(\mathbf{M}, \mathbf{y})\in \{m_{\rm st}\cup m_{\rm lt}\}}} \mathbb{E}_{(\mathbf{M}, \mathbf{y})} [\mathcal{L}_{F}(f(\mathbf{M};\theta_f), \mathbf{y})],
    \label{eq1}
\end{equation}
where $\mathbf{M}$ represents a person's image patch and $\mathbf{y}$ for label. $f$ is the feature extractor to be learned, $\theta_f$ is the parameter of $f$, and $\mathcal{L}_{F}$ is the loss function. 
In this work, inspired by \cite{somers2023body}, a representation that is robust to occlusions is learned by employing a mixed loss approach. This approach combines cross-entropy loss $\mathcal{L}_{\rm CE}$ with part triplet loss $\mathcal{L}^{\rm parts}_{\rm triplet}$, formulated as follows:
\begin{equation}
    \mathcal{L}_{F} = \sum_{i\in\{g,c\}} \mathcal{L}_{\rm CE}(h_i(\mathbf{F}), \mathbf{y}) + \mathcal{L}^{\rm parts}_{\rm triplet}(\mathbf{F}, \mathbf{y}),
\end{equation}
where $\mathbf{F}$ denotes the part features. The cross-entropy loss $\mathcal{L}_{\rm CE}$ focuses on optimizing the feature extractor to accurately predict the person's identity $\mathbf{y}$ from each holistic feature ${h_{g}, h_{c}}$. The global feature $h_{g}\in \mathbb{R}^{C}$ is obtained through global average pooling, while the concatenated feature $h_{c}\in \mathbb{R}^{(C\cdot N)}$ is derived by concatenating $N$ part features. Moreover, the part triplet loss $\mathcal{L}^{\rm parts}_{\rm triplet}$ considers a triplet comprising a query sample, the hardest positive, and the hardest negative. The hardest positive is identified as the positive sample that is furthest from the query, based on the average distance across part features, calculated as $d^{ij}_{\rm parts}=\frac{1}{N}\sum_{k=1}^{N}||\mathbf{F}_k^i-\mathbf{F}_k^j||_2$. At the same time, the hardest negative is selected as the nearest negative sample to the query.

By continually learning from these experiences ($m_{\rm lt}\cup m_{\rm st}$), the feature extractor incrementally acquires current knowledge while retaining previous experiences. This can be demonstrated by Fig.~\ref{tsne} (c), which shows that training the feature extractor in an OCL manner leads to the target person's features being distinguishable from others throughout the observed samples in the sequence. This incremental learning ability enables the robot to re-identify the target person if their re-appearance exists in the previous experiences.

After feature extraction, previous works\cite{ye2021deep,Li2014CVPR,somers2023body} usually achieve ReID by averaging the similarities of features across all query-gallery pairs, assuming that the query feature and the gallery features are strictly in the same feature space. This requires one to re-extract features with the latest feature extractor from all samples in the memory buffer. However, for the purpose of effective RPF, this approach is not feasible due to the large size of our long-term memory. Therefore, to ensure efficient ReID processing, we leverage short-term experiences to train a classifier (Sec.~\ref{sec3-3}).

\subsection{Target Classifier}\label{sec3-3}
We train a target classifier $g$ using short-term experiences $m_{\rm st}$, representing the latest knowledge about the target person. Here, we employ the ridge regression (RR) model with L2 regularization as our classifier, although any other classifiers that are capable of efficient optimization and inference can also be employed. Specifically, we train $N$ RR models where each model is represented as $\mathbf{W}_i\in{\mathbb{R}^{1\times{C}}}$ corresponding to a part-level classifier. The target confidence $s$ is estimated by averaging the outputs from all part-level classifiers:
\begin{equation}
    s = \frac{\sum_{i=1}^{N} v_i \mathbf{W}_i \mathbf{F}_i^T}{\sum_{i=1}^{N} v_i},
\label{eq3}
\end{equation}
where $v_i$ is the visible indicator and $\mathbf{F}_i\in\mathbb{R}^{1\times{C}}$ represents the $i_{th}$ part feature of $\mathbf{F}$ where $\mathbf{F}=\bar{f}(\mathbf{M})$. Each RR model $\mathbf{W}_i$ is optimized with the most recent $K$ features extracted from $m_{\rm st}$: 
\begin{equation}
    \mathop{\arg\min}_{\mathbf{W}_i} \left \|\mathbf{W}_i\mathbf{X}_i^T - \mathbf{y} \right \|^2_2 + \lambda  \left \| \mathbf{W}_i \right \|^2_2,
\label{eq4}
\end{equation}
where $\mathbf{X}_i=\{\mathbf{F}^1_i,\mathbf{F}^2_i,\ldots,\mathbf{F}^{K}_i\}\in{\mathbb{R}^{K\times{C}}}$ represents the features of the $i_{\rm th}$ part. $\mathbf{y}$ indicates the labels and $\lambda$ is a regularization parameter. The optimal solution, which is obtained using linear least squares, is given by $\mathbf{W}_i^*=(\mathbf{X}_i^T\mathbf{X}_i+\lambda{\mathbf{I}})^{-1}\mathbf{X}_i^T\mathbf{y}$.  

This formulation can efficiently regress the classification boundary since the sizes of both the short-term memory and feature dimensions are small. Furthermore, it can also generalize to distinguish historical samples, although the classifier is trained on short-term experiences only. This is because optimized features are discriminative enough to establish a clear classification boundary with a few samples. This can be observed from Fig.~\ref{tsne} (c) and further verified in the experiments.

\subsection{Memory Manager}\label{sec3-4}
To leverage long-term experiences to mitigate domain drift problems, we establish a long-term memory denoted as $\mathbb{L}$, responsible for storing valuable samples, i.e., pairs of image patches and labels. When presented with a new sample, the memory manager employs a \textit{keyframe selection} strategy to decide whether to add this sample to the memory buffer. Once the buffer reaches its capacity, \textit{memory consolidation} takes effect to create space by purging certain samples. 
In addition to the sample insertion and removal, the process of selecting samples for replay during model optimization (Eq.~\ref{eq1}) is equally important and is overseen by the \textit{memory replay} mechanism.
In the following, we will introduce our \textit{keyframe selection} strategy, as well as the \textit{memory replay and consolidation} processes.

\subsubsection{Keyframe Selection} 
Adding the newest sample directly to $\mathbb{L}$ may not be appropriate because the appearance of the target person in adjacent frames is often similar, and therefore, it may not provide additional information.
Since information in images is temporally correlated and therefore highly redundant, we insert a keyframe to $\mathbb{L}$ only if it is informative. To this end, inspired by \cite{sucar2021imap}, we employ a \textit{loss-guided} keyframe selection strategy to assess the significance of the incoming sample.
Specifically, every time a new target sample is added to $\mathbb{L}$, and the feature extractor is optimized, we save a duplicate of the latest feature extractor $f$ and record the loss from this optimization as $l_{t}$. The subsequent sample $\{\mathbf{M}, \mathbf{y}\}_{id}$ will then be used to optimize the duplicated $f$. If the optimization loss is larger than the previous loss $l_{t}$ by a margin, this sample will be added to $\mathbb{L}$. This process can be expressed as:
\begin{equation}
\delta = \mathcal{L}_F(f(\mathbf{M}),\mathbf{y}) - l_{t},
\label{eq5}
\end{equation}
if $\delta>\delta_{l}$, the sample is added to $\mathbb{L}$, indicating that the forthcoming sample contributes additional information to the learned feature extractor.

\subsubsection{Memory Replay and Consolidation}\label{memory}
To preserve valuable experiences, we follow the standard technique of memory replay in OCL to replay samples for our feature extractor learning or consolidate the memory by removing non-informative samples. Specifically, the consolidation process is triggered when $\mathbb{L}$ is full. For example, Reservoir\cite{ER} adds a sample to $\mathbb{L}$ with a probability of $|\mathbb{L}|/n$, where $|\mathbb{L}|$ is the size of the long-term memory and $n$ is the total number of observed samples. This is executed as follows: $\mathbb{L}[i]\leftarrow \{\mathbf{M},\mathbf{y}\}$ if $i<|\mathbb{L}|$, with $i={\rm randint}(0,n)$. This approach inherently reduces the likelihood of later samples being sampled. Differing from a selection based on sequential observation, BioSLAM\cite{yin2022bioslam} chooses to remove non-discriminative samples from long-term memory via online clustering. In this work, we demonstrate that by leveraging existing OCL techniques, we can effectively address the issue of forgetting and enhance the person ReID capability. Besides, the memory management module itself does not solve the domain drift problem but is axillary to the feature extractor that directly tackles the domain drift problem.

Our framework optimizes the feature extractor and the classifier upon successful identification of the target person, identification recognized when the target $id$ exists within the tracked individuals and the target confidence $s$ surpasses the threshold $\delta_{\rm sw}$. When the target person is lost, the algorithm re-identifies him from all observed individuals. An individual is considered the target person if his estimated confidence has surpassed a threshold $\delta_{\rm sw}$ for consecutive $\zeta_{\rm reid}$ frames.

\section{Experiments}
\subsection{Experimental Setup}
To ensure fair and consistent evaluation in terms of target-person-tracking ability for RPF, previous RPF works (such as \cite{chen2017integrating, koide2020monocular, ye2023icra}) typically assess person-following performance by evaluating person-tracking performance on a robot-centric following dataset.
\subsubsection{Dataset} 
We conduct experiments on a public dataset\cite{chen2017integrating} and a custom-built dataset. Both datasets consist of image sequences with the ground truth provided in the form of bounding boxes around the target person.
The public dataset includes challenging scenarios such as quick multi-people-crossing, illumination changes, and appearance variations. However, this public dataset lacks scenarios that require person ReID, such as occlusion and similar appearances of distracting people. To address this limitation, we created a custom dataset that includes these challenging scenarios. The custom dataset comprises four sequences named \textit{corridor1}, \textit{corridor2}, \textit{lab-corridor}, and \textit{room}. 

\begin{table*}[t]
    \vspace{5pt}
	\caption{Success rate of person tracking (\%) of the baseline and our method in the custom-built dataset$^\dag$ and the public dataset\cite{chen2017integrating}. Our complete RPF system achieves the highest success rate due to the effective ReID performance of our OCL-assisted RPF-ReID module.}
	\centering
	\scalebox{1.08}{
		\begin{tabular}{l|ccccc}
			\toprule
            \multirow{2}{*}{\textbf{Methods}} &\multicolumn{5}{c}{\textbf{Success Rate (\%)}} \\
            
            &\textbf{\textit{corridor1}}$^\dag$ &\textbf{\textit{corridor2}}$^\dag$ &\textbf{\textit{lab-corridor}}$^\dag$ &\textbf{\textit{room}}$^\dag$ &\textbf{\textit{public dataset}}\cite{chen2017integrating} \\ 
            
            \midrule
            \midrule
			Zhong's Method\cite{zhong2021towards} &63.8 &66.8 &75.8 &44.7 &75.8 \\
			SiamRPN++\cite{li2019siamrpn} &44.8 &55.9 &46.1 &42.6 &93.6 \\
			STARK\cite{yan2021learning} &44.3 &83.8 &73.1 &65.8 &96.5 \\

            \midrule
            SORT\cite{bewley2016simple} + RPF-ReID &67.3  &37.9 &31.1 &82.4 &96.1 \\
            OC-SORT\cite{cao2023observation} + RPF-ReID &67.3 &37.9 &31.1 &82.4 &96.1 \\
            ByteTrack\cite{zhang2022bytetrack} + RPF-ReID &69.1 &20.2 &54.2 &82.4 &96.3 \\
            \rowcolor{gray!25}
            \textbf{ByteTrack + RPF-ReID + OCL} &\textbf{93.5} &\textbf{94.9} &\textbf{96.0} &\textbf{96.8} &\textbf{97.0} \\
            \bottomrule
	\end{tabular}}
	\label{tab1}
    \vspace*{-0.10in}
\end{table*}

\subsubsection{Baselines}
To verify the effectiveness of the proposed RPF framework in terms of target-person-tracking ability, we first compare it with some popular one-stage baselines. With frame input, one-stage methods directly output the target person's bounding box (e.g., SiamRPN++\cite{li2019siamrpn} and STARK\cite{yan2021learning}) or camera movement (e.g., Zhong's Method\cite{zhong2021towards}). Moreover, we compare it with some people trackers (e.g., SORT\cite{bewley2016simple}, OC-SORT\cite{cao2023observation}, and ByteTrack\cite{zhang2022bytetrack}), which can track people's bounding boxes based on motion models. However, these trackers alone cannot handle the target-person-tracking situation and are auxiliary to a target-ReID module for re-identifying the target person after long-term occlusion. Therefore, a complete RPF system is formed by combining each of these people trackers with our ReID module. Our complete RPF system includes ByteTrack\cite{zhang2022bytetrack} and the OCL-assisted RPF-ReID module, labeled as ``\textbf{ByteTrack + RPF-ReID + OCL}'' as shown in Table~\ref{tab1}.

To investigate the impact of different methods of memory consolidation (as illustrated in Sec.~\ref{sec3-4}) on person ReID ability, we conduct experiments involving three methods: BioSLAM\cite{yin2022bioslam}, MIR\cite{Rahaf2019nips}, and Reservoir\cite{ER}. These methods are employed to assess whether any form of memory consolidation can enhance the performance of person ReID. Experimental analysis is shown in Sec.~\ref{sec:OCLExp}.

\subsection{Evaluation of our RPF system}
\subsubsection{Metric}
The evaluation metric of person tracking relies on those employed in previous RPF studies\cite{chen2017integrating, koide2020monocular, ye2023icra}. We assess tracking performance in the image space using the success rate of person tracking as the evaluation metric, which is calculated as $\frac{1}{N} \sum_{i=0}^N a_i$,
where $N$ represents the number of frames within a sequence and $a_i$ is a binary indicator. It equals one if the distance between the recognized and ground-truth bounding boxes is less than 50 pixels and zero otherwise. As for Zhong's method\cite{zhong2021towards}, which is a reinforcement-learning-based tracker outputting the action directly, we compare it in the action space (detailed settings are explained in APPENDIX-A).

\subsubsection{Experimental Results}
The results are shown in Table \ref{tab1}. It can be observed that a people tracker with our RPF-ReID can achieve better performance than one-stage methods in \textit{corridor1} and \textit{room}. For example, ``ByteTrack\cite{zhang2022bytetrack} + RPF-ReID'' achieves 69.1\% and 82.4\% respectively. However, this combination achieves a lower success rate with 20.2\% and 54.2\% in \textit{corridor2} and \textit{lab-corridor}, respectively. These two datasets contain significantly more long-term variations, with more than 5,000 frames recorded. Pre-trained deep features cannot generalize well to these practical RPF scenarios, which include different viewpoints and lighting conditions, suffering from the so-called domain drift problem.

When the feature extractor is fine-tuned online using our OCL-based memory manager (``ByteTrack + RPF-ReID + OCL''), it achieves the best performance, with success rates above 93.5\% in all sequences of the custom-built dataset and 97.0\% in the public dataset. This indicates the effectiveness of our proposed OCL-based memory manager in leveraging online collected experiences to optimize the feature extractor and mitigate domain drift. In this way, the resulting ReID features capture more knowledge about the target person, enabling successful ReID even in challenging RPF scenarios.

\subsection{Online Continual Learning Evaluation}\label{sec:OCLExp}
\subsubsection{Metric}
The evaluation of OCL\cite{mai2022online} aims to assess how well the model remembers previous knowledge, which is essential for person ReID in RPF, as previous knowledge contains potentially matching experiences for future ReID. Additionally, incrementally remembering previous knowledge might result in a more generalized feature extractor.
Specifically, we treat the OCL evaluation for person ReID as a classification task, where we assume that the true identity of the target person is known in each frame, and the model incrementally learns with known labels. For evaluation purposes, we divide each sequence into eight segments, each representing different levels of distribution drift. During incremental learning, after each segment is learned, the model is evaluated on previously seen segments. Similar to \cite{mai2022online}, we use the ReID mean accuracy at the end of training (r-mEAcc) as our OCL evaluation metric: $\frac{1}{8} \sum_{j=0}^8 a_{8,j}$,
where $a_{8,j}$ represents the average accuracy on the $j_{\rm th}$ segment, with the model learned from all eight segments. Higher r-mEAcc values indicate that the model retains more of the previous knowledge during incremental learning.

\begin{table}[t]
    \centering
\caption{Experiments on \textit{corridor2} and \textit{lab-corridor} are conducted to evaluate the ReID mean accuracy at the end of training (r-mEAcc, \%) and success rate (SR, \%). All r-mEAcc values are averaged across three runs.}
\scalebox{0.56}{
    \begin{tabular}{l|cc|cc}
        \toprule
        \multirow{2}*{\textbf{Methods}}
        
        &\multicolumn{2}{c|}{\textbf{corridor2}} &\multicolumn{2}{c}{\textbf{lab-corridor}}\\
        &\textbf{\textit{r-mEAcc} $\uparrow$} &\textbf{\textit{SR} $\uparrow$}
                                &\textbf{\textit{r-mEAcc}} &\textbf{\textit{SR}}\\
        
        \midrule
        ByteTrack\cite{zhang2022bytetrack} + RPF-ReID &59.2 $\pm$ 0.0 &20.2  &31.7 $\pm$ 0.0 &54.2\\
        ByteTrack + RPF-ReID + OCL, based on BioSLAM\cite{yin2022bioslam} &94.9 $\pm$ 2.0 &94.9  &79.0 $\pm$ 22.5 &93.8\\
        ByteTrack + RPF-ReID + OCL, based on MIR\cite{Rahaf2019nips} &94.7 $\pm$ 0.8 &95.4  &86.1 $\pm$ 14.3 &96.1\\
        ByteTrack + RPF-ReID + OCL, based on Reservoir\cite{ER} &96.5 $\pm$ 0.4 &94.9  &94.0 $\pm$ 0.7 &96.0\\
        \bottomrule
    \end{tabular}
    }
    \label{tab-oclreid}
    \vspace*{-0.2in}
\end{table}

\subsubsection{Experimental Results}
The results are shown in Table~\ref{tab-oclreid}. The row of ``ByteTrack\cite{zhang2022bytetrack} + RPF-ReID'' utilizes a pre-trained and fixed feature extractor. In comparison to the original setup ``ByteTrack + RPF-ReID + OCL, based on Reservoir\cite{ER}'', it experiences a significant decline in performance, with reductions of 37.3\% and 74.7\% in r-mEAcc and success rate, respectively, on \textit{corridor2}. 
This underscores the importance of online optimization of the feature extractor with collected experiences to combat domain drift and enhance the system's ReID performance. Moreover, this approach to discriminative ReID modeling markedly improves person tracking efficacy, as evidenced by higher success rate.

\begin{figure}[t]
    \centering
    \begin{subfigure}[]{0.47\linewidth}
            \centering
            \includegraphics[width=\linewidth]{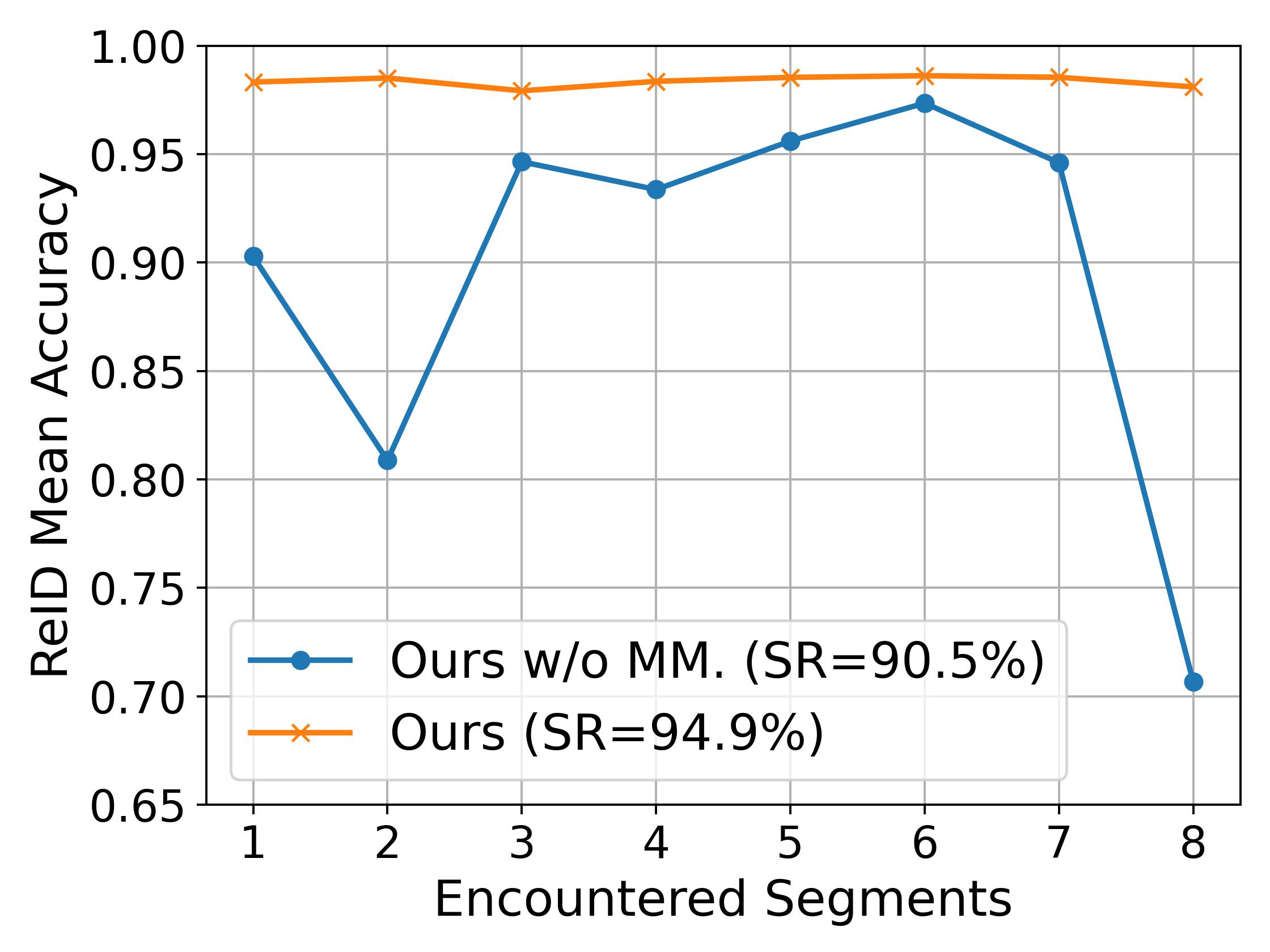}
            \caption{corridor2}
            \label{accCorridor2}
    \end{subfigure}%
    \hspace{0.01\linewidth}
    \begin{subfigure}[]{0.47\linewidth}
            \centering
            \includegraphics[width=\linewidth]{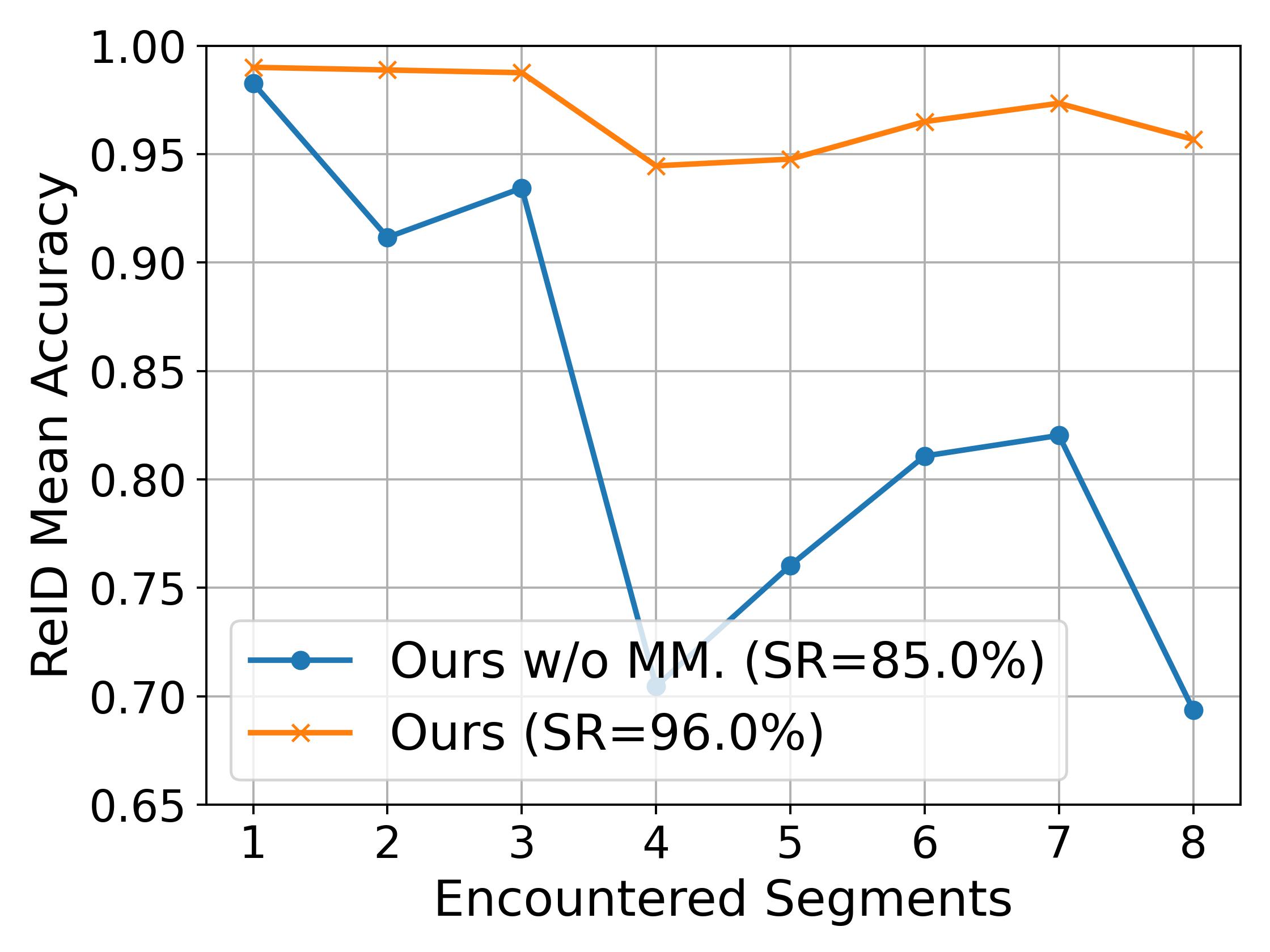}
            \caption{lab-corridor}
            \label{accLabCorridor}                           
    \end{subfigure}%
\caption{Plots of ReID mean accuracy w.r.t. encountered segments. ``Ours w/o MM.'' indicates fine-tuning the feature extractor without memory management (introduced in Sec.~\ref{sec3-4}), using newly observed samples only. After fine-tuning from a new segment, the model is evaluated on segments it has encountered previously to determine its mean accuracy. This metric indicates the model's ability to retain knowledge from segments learned earlier.}
\label{accs}
\vspace*{-0.20in}
\end{figure}

We also verify the necessity of our memory management (introduced in Sec.~\ref{sec3-4}) for mitigating domain drift. We use newly observed samples only to fine-tune the feature extractor without memory management. As shown in Fig.~\ref{accs}, this naive strategy (``Ours w/o MM.'') leads to significant domain drift, resulting in a notable decrease in ReID accuracy across different segments. Such drift significantly undermines the RPF system's tracking efficiency, manifesting as success rate reductions of 4.4\% and 11.0\% on \textit{lab-corridor} and \textit{corridor2}, respectively. This outcome highlights the value of our memory manager in preserving valuable long-term experiences. By replaying these experiences to mitigate domain drift, we ensure that our ReID features remain robust, thereby enhancing the RPF system's consistent tracking performance.

In summary, the above experiments demonstrate the effectiveness of optimizing the feature extractor online using collected experiences managed by our memory manager. This strategy effectively addresses domain drift, resulting in enhanced ReID performance within the RPF system. Another observation is that although the OCL ability of BioSLAM\cite{yin2022bioslam} is worse than Reservoir\cite{ER} with r-mEAcc of 79.0\% vs. 94.0\% on \textit{lab-corridor}, its tracking accuracy only drops by 2.2\%. This indicates that not all historical knowledge needs to be memorized for person ReID in some situations. However, we claim that maximizing the enhancement of ReID ability at a long-term scale is still necessary as it ensures a discriminative appearance representation for dealing with complex ReID situations.

\subsection{Runtime Analysis}
We conducted a runtime analysis on two computer setups: a high-end PC and an onboard NUC. For our proposed ReID module, we tested the following configuration, which showed the best performance in our experiments: Reservoir-based\cite{ER} memory consolidation, a ResNet18-based feature extractor where only the layers after \textit{conv3} (including \textit{conv3}) are trainable, and the RR-based part classifiers.

In the experiment, we ran a separate thread that encompasses memory management and feature extractor fine-tuning, operating independently from the main thread. The results are shown in Table~\ref{runtimeAnalysis}. On the high-end PC, the main thread runs at 35.1 Hz, and the separate thread runs at 22.2 Hz. On the onboard NUC, the threads run at 18.8 Hz and 6.6 Hz, respectively. Therefore, we conclude that our RPF system can follow a target in real time, as also demonstrated in the supplementary video.

\subsection{Implementation Details}
For all experiments, we set the following default parameters:  memory sizes $|\mathbb{S}|=64$ and $|\mathbb{L}|=512$, a batch size of $64$ for each replay including long-term and short-term relays, a regularization parameter $\lambda=1.0$ for RR, a keyframe selection threshold $\delta_{l}=0.02$, an id switch threshold $\delta_{\rm sw}=0.35$, a ReID threshold $\delta_{\rm reid}=0.7$  and a number of consecutive frames $\zeta_{\rm reid}=5$. 
In this paper, for representing the part-level features, we define ten parts: \{front, back\}$\times$\{head, torso, legs, feet, whole\}.

For orientation estimation, we employ MonoLoco\cite{Bertoni2019ICCV} to infer the orientation using detected joint positions from AlphaPose\cite{alphaPose}. These joint positions are also utilized to estimate the visible parts. We utilize YOLOX-S\cite{ge2021yolox} for bounding-box detection. For the tracking module, we utilize ByteTrack as our tracking method. For our proposed ReID module, we use ResNet18 as our feature extractor, pre-trained on the MOT16 dataset\cite{milan2016mot16}. During OCL for the ResNet18, only the layers after \textit{conv3} are trainable (including \textit{conv3}). 

All evaluations are conducted on both a high-end PC and an onboard NUC. The high-end PC includes an Intel® Core™ i9-12900K CPU and NVIDIA GeForce RTX 3090. The onboard NUC is an Intel NUC 11 mini PC powered by a Core i7-1165G7 CPU and NVIDIA GeForce RTX2060-laptop GPU. This NUC is mounted on a Unitree Go1 quadruped robot to perform robot person following in the real world as shown in Fig.~\ref{introduction} and the submitted video. Besides the computer, a dual-fisheye Ricoh camera is mounted on the robot, providing cropped perspective images with a resolution of $640\times 480$ and a frequency of 30Hz.

\begin{table}[t]
    \centering
    \caption{Runtime analysis of our RPF framework. Two computer setups are evaluated, including a high-end PC (\textbf{Setup 1}) and an onboard NUC (\textbf{Setup 2 for real-world deployment}). We record the average time cost (\textbf{ms}) per frame. \textbf{Perception} includes bounding-box, human-joint and human-orientation detections. \textbf{Tracking} denotes the motion tracker. \textbf{ReID} indicates the re-identification process, including feature extraction and target estimation. The above three processes run in the \textbf{Main Thread}. A \textbf{Separate Thread} handles the OCL process, including memory management and replay, as well as the online continual learning of the feature extractor. For detailed experimental settings, refer to the supplementary materials.}
    \scalebox{0.68}{
        \begin{tabular}{c|cccc|c}
            \toprule
            \multirow{2}*{\textbf{Setup}} &\multirow{2}*{\textbf{Perception}} &\multirow{2}*{\textbf{Tracking}} &\multirow{2}*{\textbf{ReID}} &\textbf{Total} &\textbf{OCL}\\
            & & & &\textbf{(Main Thread)} &\textbf{(Separate Thread)}\\
            \midrule
            \midrule
            1 &18.1 &1.7 &8.7 &28.5 &45.1\\
            \rowcolor{gray!25}
            2 &33.4 &2.4 &12.9 &53.2 &152.0\\
            \bottomrule
        \end{tabular}
    }
    \label{runtimeAnalysis}
    \vspace*{-0.1in}
\end{table}

\section{CONCLUSION} \label{sec:conclusion}
We approach person ReID in RPF as a problem of online continual learning for mitigating the domain drift problems. This enables the RPF system to learn incrementally from online collected experiences.
As a result, the framework achieves a discriminative appearance representation, allowing for effective ReID even in challenging scenarios, such as frequent appearance changes, occlusion, and distracting people with similar appearances. Compared to existing baselines, our target-ReID framework achieves state-of-the-art performance in person ReID within RPF scenarios.

For future work, i) we will explore methods to consolidate valuable samples, aiming to maximize the learning of appearance representations while preventing the forgetting of previous knowledge. Additionally, strategies for balancing efficient ReID with incremental memorization in crowded environments will be investigated. ii) We will create an application-driven dataset containing practical RPF scenarios to support the development of person-tracking algorithms for RPF. These efforts will enhance the robustness and effectiveness of the OCLReID framework in real-world applications. Examples include a trolley-cart following system in airports \cite{xie2024autonomous} and shopping-cart assistance \cite{doering2015user}, which is designed to aid elderly individuals in dynamic environments.





\bibliographystyle{./bibliography/IEEEtran}
\bibliography{./bibliography/ref}
\clearpage
\onecolumn
\section*{APPENDIX}
\subsection{Comparison with Active Object Tracking Methods}

\begin{figure}[ht]
    \centering
    \includegraphics[width=0.58\linewidth]{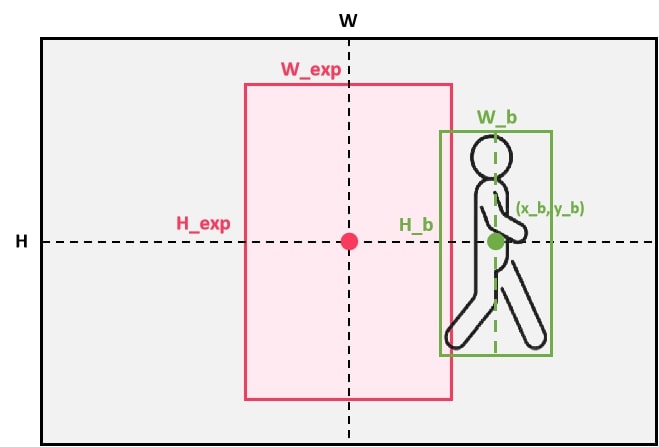}
    \caption{An example to illustrate errors is as follows: The goal of a correct action is to move the target person closer to the center of the image. Here, $(W,H)$ represents the image's width and height. $(W_{exp}, H_{exp})$ denotes the size of the expected centered bounding box, initialized by the first bounding box of the target person. $(W_b,H_b)$ represents the target person's bounding box size in the current frame, and $(x_b,y_b)$ is its center point.}
    \label{fig:bbox2action}
\end{figure}

Active Object Tracking (AOT) methods, as described in \cite{zhong2021towards}, utilize an end-to-end approach through reinforcement learning. These methods process raw video frames as input and generate camera movement actions as output. According to \cite{zhong2021towards}, there are seven discrete actions: \textit{move-forward/backward, turn-left/right, move-forward-and-turn-left/right}, and \textit{no-op}. However, existing person-following datasets, such as Chen's dataset \cite{chen2017integrating} and our own dataset, only provide ground truth bounding boxes of the target person. For a valid comparison, we need to map these bounding boxes to the action space. As noted in \cite{zhong2021towards}, a two-stage method (combining single object tracking with a PID controller) can achieve a 100\% success rate if the estimated target bounding box (bbox) is accurate. Therefore, we can deduce the ground truth actions from the ground truth target bboxes. This approach allows us to evaluate AOT methods within person-following datasets. According to \cite{zhong2021towards}, the objective of the camera action is to keep the target person's bbox centered in the image, maintaining the same size as initially observed. The ground truth action is generated according to the horizontal error $X_{err}=\frac{X_b-W/2}{W/2}$ and the size error $S_{err}=\frac{W_b\times H_b - W_{exp}\times H_{exp}}{W_{exp}\times H_{exp}}$ shown as in Fig.~\ref{fig:bbox2action}. According to \cite{zhong2021towards}, we have the following ground truth action mappings:
\begin{enumerate}[(1)]
    \item \textit{Move forward} if ${\rm abs}(X_{err})\leq0.1$ and $S_{err}\leq-0.2$;
    \item \textit{Move backward} if ${\rm abs}(X_{err})\leq0.1$ and $S_{err}\geqslant0.2$;
    \item \textit{No-op} if ${\rm abs}(X_{err})<0.1$ and ${\rm abs}(S_{err})<0.2$;
    \item \textit{Move forward and turn right} if $0.1 \leq X_{err}\leq 0.3$;
    \item \textit{Turn right} if $X_{err}>0.3$;
    \item \textit{Move forward and turn left} if $-0.3 \leq X_{err}\leq -0.1$;
    \item \textit{Turn left} if $X_{err}<-0.3$.
\end{enumerate}

\begin{figure}[t]
    \centering
    \begin{subfigure}[]{\linewidth}
        \centering
        \includegraphics[width=\linewidth]{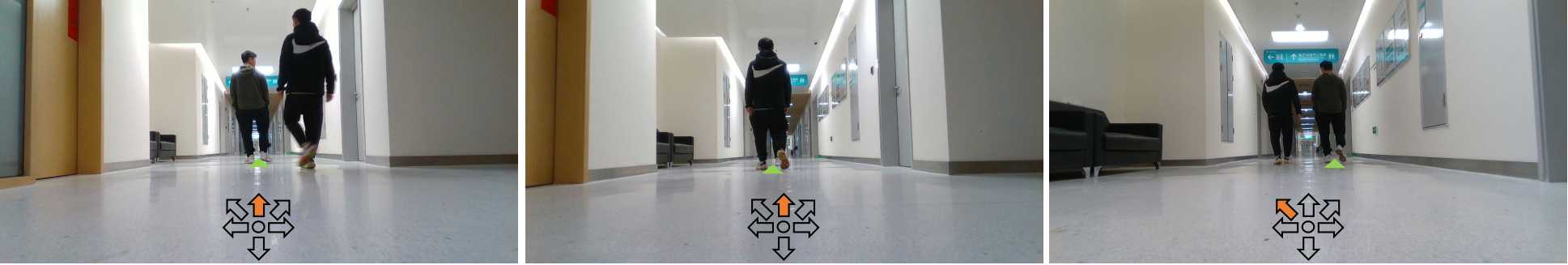}
        \caption{Zhong's method\cite{zhong2021towards}}
        \label{rlVis}
    \end{subfigure}%
    \\
    \hspace{0.1\linewidth}
    \begin{subfigure}[]{\linewidth}
        \centering
        \includegraphics[width=\linewidth]{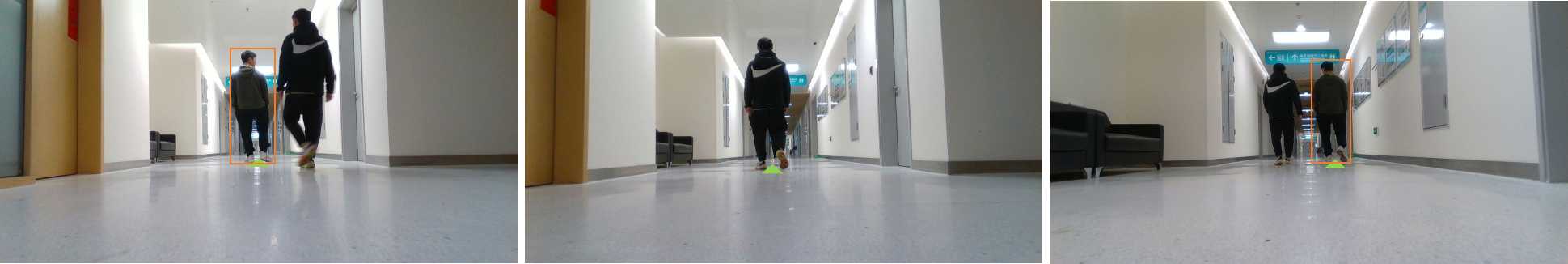}
        \caption{Ours}
        \label{oursVis}
    \end{subfigure}%
\caption{An example comparing Zhong's method \cite{zhong2021towards} and ours. From left to right, the sequence represents observations where a long-term occlusion occurs. (a) Zhong's method outputs \textit{move forward and turn left} due to a failed re-identification of the target person. (b) Our method reliably re-identifies the target person even after long-term occlusion.}
\label{fig:rlAndOurs}
\end{figure}

This paper aims to evaluate the algorithms' tracking performance, meaning true target person identification is the first priority. Therefore, besides evaluating accurate action estimation, we reduce the matching standards. Specifically, we consider an action to be true if this action tries to move the target person to the center of the image. For example, if the bbox of the target person is on the left of the image, \textit{move backward, turn left}, and \textit{move forward and turn left} are all considered as true actions. The results are reported in Table.~\ref{tab1}. An example is shown in Fig.~\ref{fig:rlAndOurs}. We observe that after a long occlusion by a visually similar person, Zhong's method outputs \textit{move forward and turn left}, failing to re-identify the target person. In contrast, our method reliably re-identifies the target person even after long-term occlusion.
\clearpage

\subsection{Memory Examples}
\begin{figure}[ht]
    \centering
    \includegraphics[width=0.58\linewidth]{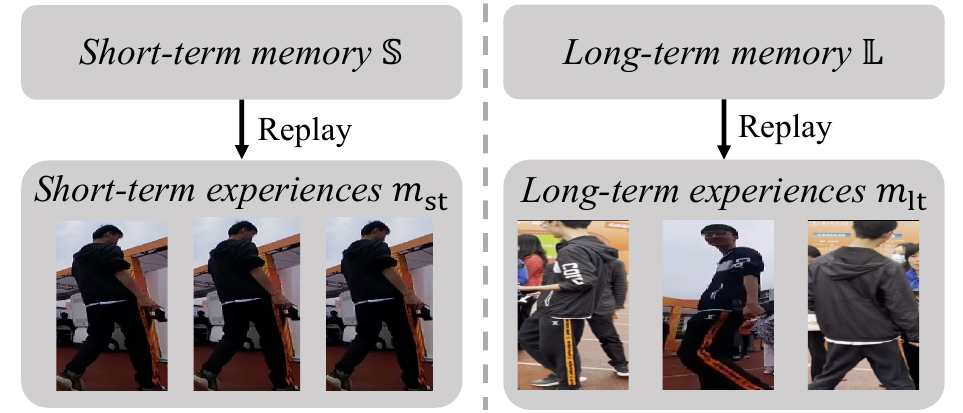}
    \caption{Both short-term and long-term experiences $(m_{\rm st} \cup m_{\rm lt})$ are responsible for training the feature extractor with $m_{\rm st}$ and $m_{\rm lt}$ being sampled from short-term memory $\mathbb{S}$ and long-term memory $\mathbb{L}$, respectively. $\mathbb{L}$ contains sparse yet valuable historical samples, maintained by the \textit{memory manager}. Besides, the target classifier is trained with $m_{\rm st}$ sampled from $\mathbb{S}$, which stores the most recently observed samples, representing the latest knowledge.}
    \label{fig:memoryExamples}
\end{figure}

Several replayed examples of our short-term and long-term memories are shown in Fig.~\ref{fig:memoryExamples}. For more visual examples of short-term and long-term memories, please refer to the supplementary video.

\subsection{Visual Examples during RPF Task}
\begin{figure}[t]
    \centering
    \begin{subfigure}[]{\linewidth}
        \centering
        \includegraphics[width=0.55\linewidth]{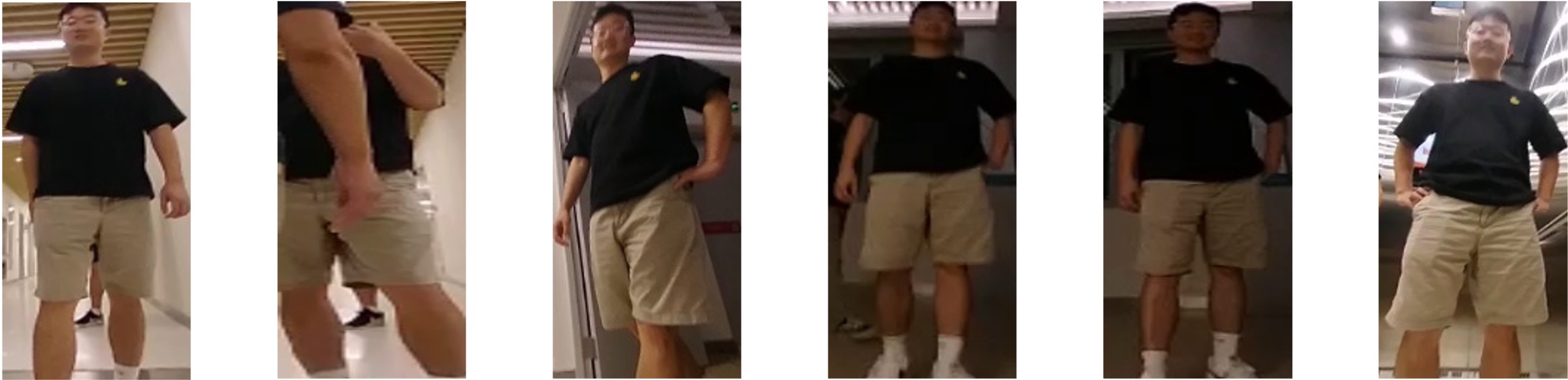}
        \caption{Lighting changes}
        \label{fig:datasetExamples:lighting}
    \end{subfigure}%
    \\
    \begin{subfigure}[]{\linewidth}
        \centering
        \includegraphics[width=0.55\linewidth]{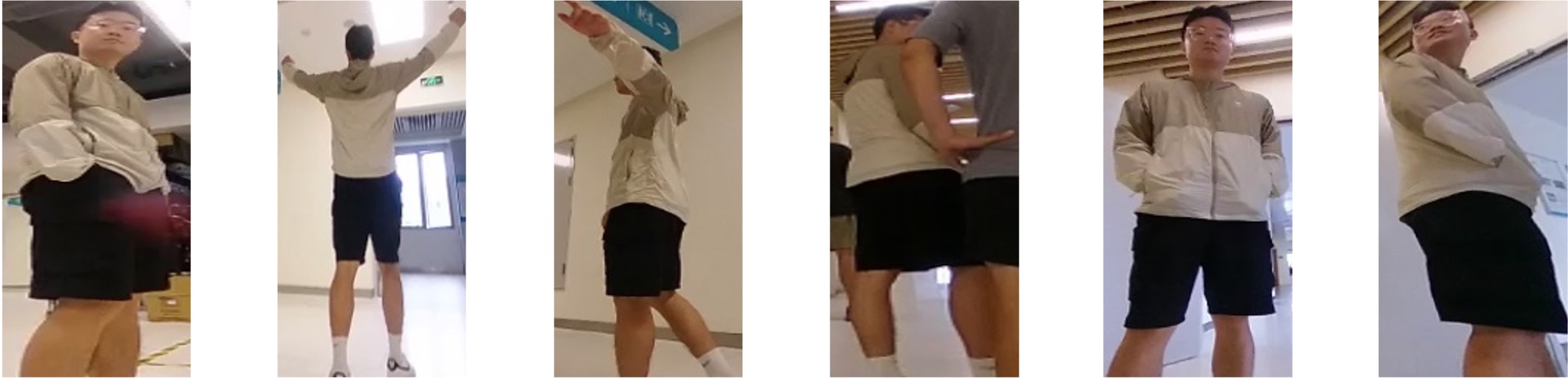}
        \caption{Viewpoint changes}
        \label{fig:datasetExamples:viewpoint}
    \end{subfigure}%
\caption{Sampled images of the target person during the robot-person-following task. We can observe significant changes in the appearance of the target person from left to right. (a) Lighting changes: bright light in the corridor, dim light in a corner, and bright light again in the elevator. (b) Viewpoint changes: the person's front view, back view, occlusion, and front view again.}
\label{fig:datasetExamples}
\end{figure}

Sampled images are shown in Fig.~\ref{fig:datasetExamples} consisting of the observation of the target person during the robot-person-following task. We can observe significant changes in the appearance of the target person from left to right including lighting and viewpoint changes. In such situations, previous methods relying short-term experiences would fail to re-identify the target person after occlusion. For example, as shown in Fig.~\ref{fig:datasetExamples} (b), short-term experiences capture the latest observation, i.e., the back view of the target person. Consequently, when the target person reappears with a front view, these short-term experiences fail to re-identify them.

In contrast, our method utilizes both short-term and long-term experiences. Specifically, we fine-tune the feature extractor within our OCL-assisted RPF-ReID module using both types of experiences. This approach constructs a complete representation of the target person, considering long-term experiences (e.g., the target person's front view) and short-term experiences (e.g., the target person's back view). As a result, our method can re-identify the target person even when they reappear with a front view.
\clearpage

\subsection{Target-ReID Lifecycle}
\begin{algorithm}[ht]
    \DontPrintSemicolon
    \SetNoFillComment
    \footnotesize
    \KwIn{Current image $\mathbf{I}$ and tracked people $\{\mathbf{B}, \mathbf{p}\}_{i}$ representing bounding boxes and positions, target person's identity $id$, target confidence $s$, short-term memory $\mathbb{S}$, long-term memory $\mathbb{L}$, feature extractor $f$ and target classifier $g$}
    \KwOut{Target person's position $\{\mathbf{p}\}_{id}$ in the current frame}

    Extract image patches $\mathbf{M}$ from $\mathbf{I}$ and $\mathbf{B}$;\\
    Construct the observation set $\{\mathbf{M}, \mathbf{y}\}_{i}$ where $\mathbf{y}=1$ if $i==id$, otherwise $\mathbf{y}=0$;\\
    Extract features $\mathbf{F}$ from $\mathbf{M}$ with $f$;\\
    \uIf {$id\in\{i\}$} {
        Estimate $s$ of the target person based on Eq.~3;\\
        \uIf {$s>\delta_{\rm sw}$}{
            Consider $\{\bar{i}\}$ as identities of negative tracks;\\
            $\{\mathbf{M}, \mathbf{y}\}_{id} \rightarrow \mathbb{S}$, $\{\mathbf{M}, \mathbf{y}\}_{\bar{i}} \rightarrow \mathbb{S}$ based on FILO rule;\\
            Sample $m_{\rm st}$ from $\mathbb{S}$; \\
            Train $g$ with $m_{\rm st}$ based on Eq.~4;\\
            \textcolor{blue}{\#\#\# Separate Thread \#\#\#} \\
            $\{\mathbf{M}, \mathbf{y}\}_{\bar{i}} \rightarrow \mathbb{L}$ based on FILO rule;\\
            $\{\mathbf{M}, \mathbf{y}\}_{id} \rightarrow \mathbb{L}$ if it is a keyframe based on Eq.~5;\\
            Consolidate $\mathbb{L}$ with OCL techniques if $\mathbb{L}$ is full;\\
            Sample $m_{\rm lt}$ from $\mathbb{L}$;\\
            Train $f$ with $m_{\rm st}$ and $m_{\rm lt}$ based on Eq.~1;\\
            \textcolor{blue}{\#\#\# Separate Thread \#\#\#} \\
            \textbf{Return} target position $\{\mathbf{p}\}_{id}$;
        } \Else {
            Let $id=-1$, indicates id switch between the target person and other people;\\
            \textbf{Return} \o;
        }
    } \Else {
        Estimate $s$ of the $i_{th}$ person based on Eq.~3;\\
        \uIf {$s>\delta_{\rm reid}$ \rm for consecutive $\zeta_{\rm reid}$ frames}{
            Let $id=i$, indicates successful target person ReID;\\
            \textbf{Return} target person's position $\{\mathbf{p}\}_{id}$;
        } \Else {
            \textbf{Return} \o;
        }
    }
\caption{Target-ReID Lifecycle}
\label{identification}
\end{algorithm}

\end{document}